\documentclass[10pt,journal,compsoc]{IEEEtran}
{}
{}
{}

\usepackage{algorithm}
\usepackage{algpseudocode}
\usepackage{pifont}
\usepackage{graphicx}
\usepackage{subcaption}

\usepackage{array}
\newcolumntype{M}[1]{>{\centering\arraybackslash}m{#1}}

\ifCLASSOPTIONcompsoc
  \usepackage[nocompress]{cite}
\else
  \usepackage{cite}
\fi
\ifCLASSINFOpdf
\else
\fi
\begin{document}
%
\title{Palmprint image registration using convolutional neural networks and Hough transform}
%
%
%
%

\author{Mohsen~Ahmadi, 
        Hossein~Soleimani
}

\IEEEtitleabstractindextext{%
\begin{abstract}
Minutia-based palmprint recognition systems got lots of interest in the last two decades. Due to a large number of minutiae in a palmprint, approximately 1000 minutiae, the matching process is time-consuming which makes it unpractical for real-time applications. One way to address this issue is aligning all palmprint images to a reference image and bringing them to a same coordinate system. Bringing all palmprint images to a same coordinate system, results 
 in fewer computations  during  minutia  matching.  Using convolutional neural network (CNN) and generalized 
 Hough transform (GHT), we propose a new method to register, align, palmprint images accurately. This method, finds the corresponding rotation and displacement (in both $x$ and $y$ direction) between the palmprint and a reference image. Proposed method is capable of distinguishing between left and right palmprint automatically which helps to speed up the matching process. Furthermore, designed structure of
 CNN in the registration stage gives us the segmented palmprint image which is a pre-processing step for minutia extraction. The proposed registration method followed by minutia-cylinder code (MCC) matching algorithm has been evaluated on the THUPALMLAB database, and the results show the superiority of our algorithm over most of the state-of-the-art algorithms.

\end{abstract}

}

\maketitle

\IEEEdisplaynontitleabstractindextext

%
\IEEEpeerreviewmaketitle

\IEEEraisesectionheading{\section{Introduction}\label{sec:introduction}}

%
%
%
%
\IEEEPARstart{B}{iometrics} is used to recognize or verify human identity based on  physical or behavioral characteristics. Biometric features such as face, iris, fingerprint, hand geometry, palmprint and signature have been used for human identification and recognition \cite{ZhenhuaGuoetal2009}.  Among all these features, recently, palmprint recognition has gained considerable attention as a reliable personal identification technique. Palmprint inherits many of the  fingerprint  features. Both of them are represented by the information presented in a friction ridge impression including ridge flow, ridge characteristics, and ridge structure. Due to their uniqueness and permanence, palmprint and fingerprint identification has been generally trusted \cite{Maltonietal2009}, \cite{Kongetal2009}.

In palmprint recognition systems mainly  two different types of images: 1) low resolution, i.e. less than $200$ dpi, and 2) high resolution ($500$ dpi) images are used \cite{Kongetal2009},   \cite{Zhangetal2012}. Features in the low resolution palmprints includes principal lines, wrinkles and ridges. These features define the palmprint as a texture containing discriminatory features which are relatively stable and applicable in biometric identification or verification systems \cite{Huangetal2008}, \cite{Jiaetal2008}, \cite{WuZhao2015}. In low-resolution palmprint images,  ridges are not seen, and it is not feasible to extract the second-level information, which is more discriminant \cite{DaiZhou2011}.  Features obtained from low resolution images are more suitable for civil and commercial application such as access control \cite{Kongetal2009}. On the other hand, features obtained from high resolution images are minutiae, the orientation image, singular points, and the ridge frequency image. High resolution images are suitable for forensic application such as criminal detection \cite{Mangold2016}. Because these features are robust against time passing, they are considered admissible in a court of law \cite{Ashbaugh1999}. In this paper, our focus will be on high resolution palmprint images and minutia-based palmprint matching. Our  designed registration method is based on first-level information (mainly orientation image of palmprint), and  it is applicable as long as the palmprint images are high resolution .\\

In literature, a lot of effort has been made regarding minutia-based palmprint matching in order to reduce the matching time and to address other challenges. In \cite{JainFeng2008},  similarity between two palmprints is calculated by means of a weighted sum of minutiae and orientation field matching scores. Li et.al 
describes a palmprint matching algorithm where minutiae are  compared by means of a local and global matching \cite{LiShi2008}. Multi-feature-based palmprint recognition system is another strategy which is  suggested in \cite{DaiZhou2011}.
Minutia, orientation image, density map, and major creases are  extracted to achieve higher matching accuracy.    In \cite{Liuetal2013}, a new coarse-to-fine palmprint algorithm is proposed  to deal with the large number of minutiae in palmprints.  A clustering algorithm is applied to separate minutiae into different groups and make the matching process much faster. Raffaele and his colleagues, in \cite{Cappellietal2012}, proposed a  matching algorithm which is robust against skin distortion. They apply two stage minutia-based matching: 1) local minutia matching using Minutia Cylinder-Code (MCC) algorithm \cite{Cappellietal2010} and 2) new global score computation.\par

Use of spectral minutiae representation and segmenting palmprint into three different regions is another approach which aims to design faster matching algorithm\cite{Wangetal2014}. In \cite{ChenGuo2016}  ridge distance and conventional minutia descriptors are used together to speed up the matching. The difference of ridge
distance near the minutiae is calculated, and only the pairs with similar ridge distance are considered. Hence, the number of minutiae comparison is greatly reduced which results in less time complexity. In \cite{tariq2017massively}, a novel region-quality based minutia extraction algorithm is proposed to improve the accuracy of palmprint recognition. Also, authors proposed an efficient and minutiae based encoding and matching algorithm  to accelerate the palmprint matching.

Despite the available algorithms, some problems with the
palmprint matching still need to be solved for large-scale
applications \cite{Daietal2012}: skin distortion, diversity of different palm
regions, and computational complexity. Registering palmprint images into the same coordinate, has been used to address the skin distortion and problem of large number of minutia comparison during the match process \cite{Daietal2012}, \cite{SoleimaniAhmadi2018}. By registration, it does not need to compare all minutia from two pamprint. Therefore we  need to apply the matching algorithm only to the same parts of the two palmprints. This helps to mitigate the effects of skin distortion and improve matching speed. However, the registration methods proposed in \cite{Daietal2012}, \cite{SoleimaniAhmadi2018} are not accurate enough, and also for algorithm in  \cite{Daietal2012} there is no criterion to show whether the registration is successful.

To increase the matching speed and accuracy, motivated by
\cite{Daietal2012} and our previous work \cite{SoleimaniAhmadi2018}, we developed a new palmprint registration method. In this method we combine CNN and GHT to register all palmprint images with a reference image. Using CNN, we first roughly find the rotation difference between the current pamprint and the reference palmprint. Then, we find the exact rotation difference and translation between these two images by GHT. Two new criteria are used to measure the confidence of the registration. Also, using these two criteria our method is capable of recognizing left palm from right palm which can double the matching speed if we do not know whether the current palm is coming from the left hand or right hand. Furthermore, the designed structure of CNN in registration stage, gives us the segmented palmprint image from background which is a pre-processing step for minutia extraction.

Our minutiae matching stage is based on the MCC method with
some modification \cite{Cappellietal2010}, which includes a local matching stage
based on MCC, followed by a relaxation procedure to compute a
global matching score.

By proposed method, the matching time is much better than the state-of-the-art. On the other hand, because of accurate registration, we could subside the distortion effect, and as a result the matching accuracy improved. The experimental results on THUPALMLAB database \cite{Dai2012} indicate that the proposed system achieves a false non-match rate (FNMR) of $0.24\%$ while the false match rate (FMR) is controlled at $0$. As for the speed of our system, the average time of full palmprint matching is  about $6$ milliseconds.
The rest of this paper is organized as follows: Section \ref{reg} outlines the details of the proposed palmprint registration process. In section \ref{match}  matching strategy is explained. In Section \ref{exres}, the experimental results are presented and analyzed. Finally, we end up with conclusions in section \ref{conc}.


\section{Proposed palmprint Registration}\label{reg}

Most of the raw palmprint images are not in the same coordinate system. By applying registration and bringing images to same coordinate system, the matching process will be facilitated.  By registering, we  need to compare only minutiae in both  the query and gallery image which have almost the same location and direction, and this accelerates the matching process. \par
Most of the palmprint registration methods are based on intervals between fingers \cite{Wangetal2007}, \cite{Funadaetal1998} or hand contour and principle lines \cite{Han2004}. These features are suitable for low resolution images which are captured by contact-less devices,  ensuring the whole palm region and the finger roots are visible.
However, in high resolution images fingers are not seen or hand contour is incomplete and unreliable.  Orientation fields of different palms are quite similar, but the orientation of different
palm regions  is distinctive. This  makes the orientation field to be a reliable feature for palmprint registration.\par

\begin{figure}
    \centering
    \includegraphics[width=3.5in]{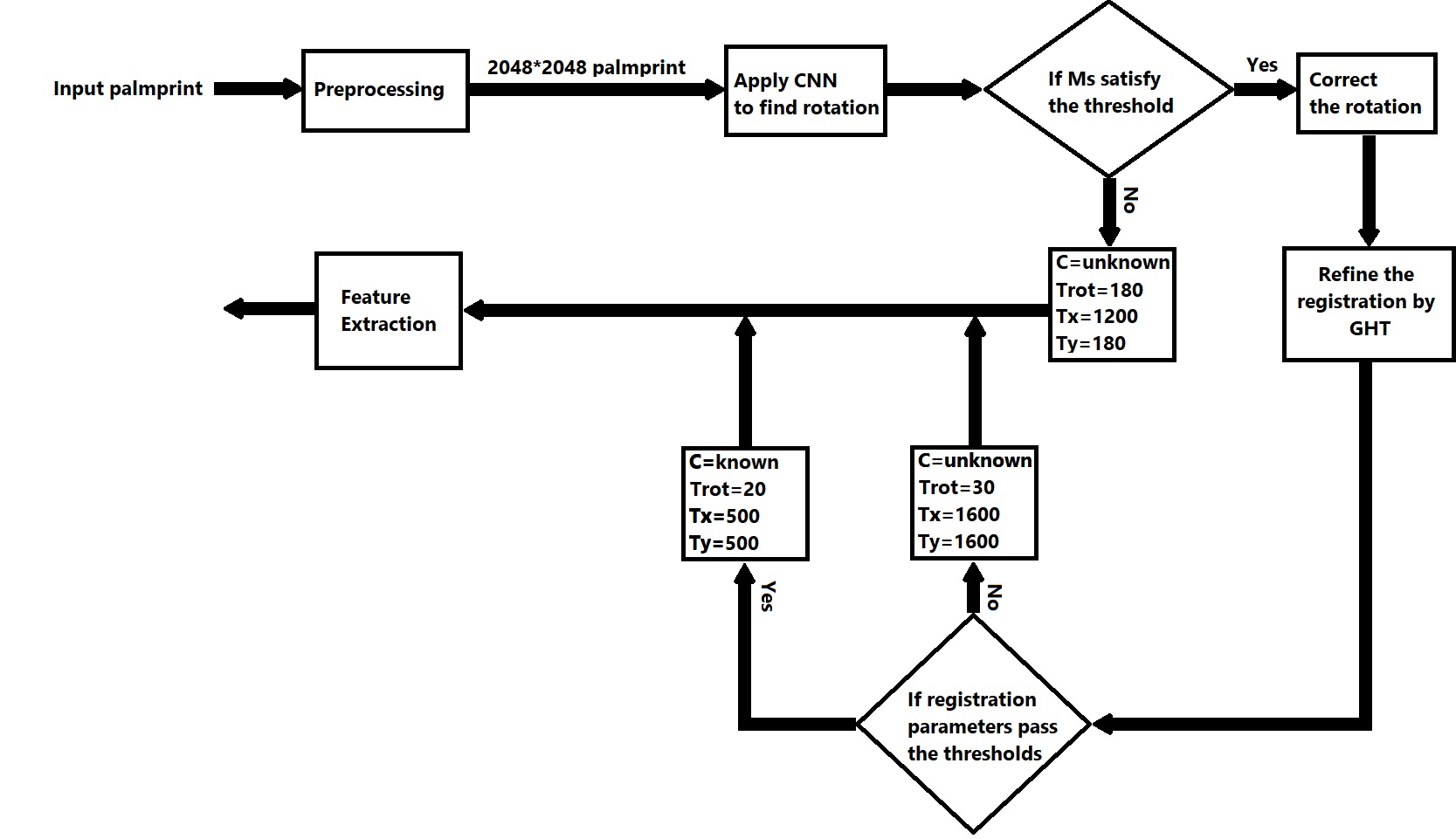}
    \caption{Steps of proposed palmprint registration method}
    \label{fig_work_flow}
\end{figure}

In \cite{Daietal2012}, \cite{SoleimaniAhmadi2018}, the average orientation field of palmprints has been used as  the reference image, and other palmprints are aligned with this image.  Parameters (rotation around $z$ axis and displacement in $x$ and $y$ direction) of rigid registration between reference orientation field and the estimated orientation field of each palmprint are obtained. This transformation (rigid body) is applied to unregistered images and brings them to the same coordinate system. These methods fail when the rotation difference between the reference image and the floating image is large (greater than $90$ degrees). The situation is much worse when the palmprint image is not that high quality, and as a result the estimated orientation field is not reliable.  To address this issue, we  first roughly find the rotation difference between two images by CNN. CNN does not need to extract orientation field, and it is robust against distortion and low quality. After that, the exact values of registration parameters is obtained by use of GHT. Figure \ref{fig_work_flow} shows the steps of proposed registration method.
 
\subsection{Registration by CNN}\label{cnn_reg}
\begin{figure}
    \centering
    \includegraphics[width=3.5in]{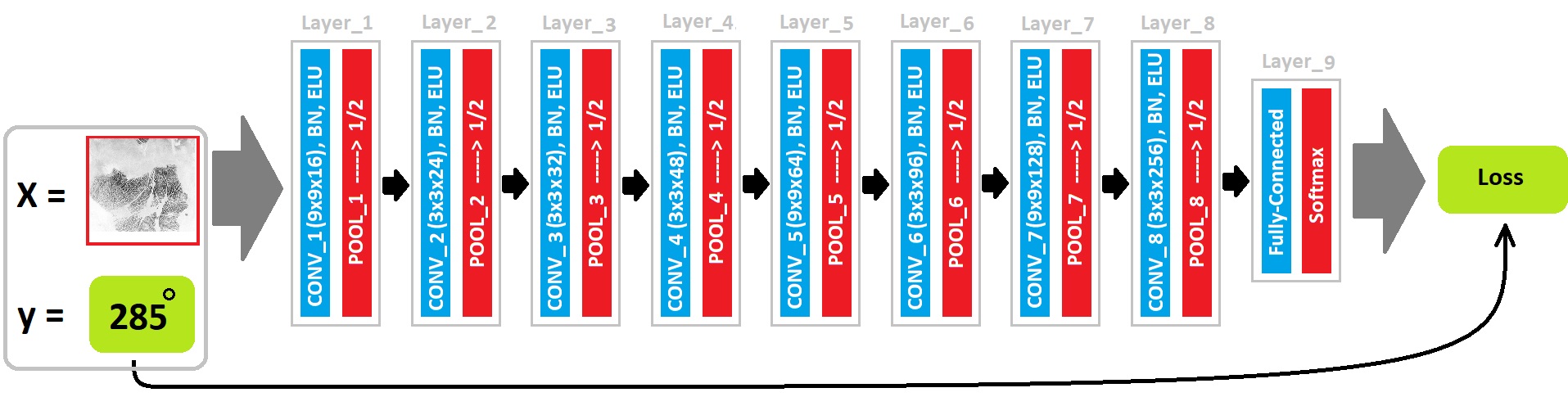}
    \caption{Architecture of designed CNN}
    \label{fig_CNN_archit}
\end{figure}

To find the rotation  of a plamprint image around $z$ axis, CNN is applied. Raw palmprint image is fed to CNN, and the output will show us the rotation of the image around $z$ axis. We formulated this task as a classification problem, i.e., the input palmprint will be classified to one of the $24$ existing classes. Theses classes are $0^\circ, 15^\circ, ..., 345^\circ$. The architecture of designed CNN is shown in figure \ref{fig_CNN_archit}.

It consists of $8$ convolutional layers, each layer includes convolution and max-pooling steps, followed by a fully connected layer and a soft-max function to calculate the final probabilities of each class. Kernel size in first layer is $9\times9$ and it changes to $3\times3$ in other convolutional layers. In the proposed network, as illustrated in  figure \ref{fig_CNN_archit}, $16$, $24$, $32$, $48$, $64$, $96$, $128$, and $256$ kernels are used for Conv layer $1$, $2$, $3$, $4$, $5$, $6$, $7$ and $8$ respectively, with stride $1$. For pooling layers, kernel size is set to $3\times3$, and stride $2$. We also applied dropout , Batch-Normalization \cite{LoffeSzegedy2015} and extensive data augmentation to overcome over fitting. Rectified Linear Unit (Relu) activation function is widely used in DNN. Because of  Dying Relu problem, we applied exponential linear unit (ELU) \cite{Clevert2016} activation function  instead of Relu  function. 

 We used  $500$ palm images in training phase, $300$ palms randomly selected from THUPALMLAB database \cite{Dai2012} and $200$ palms acquired by a live scanner. All these $500$ palms were rotated around $z$ axis such that their rotation is almost $0^\circ$ rspect to $y$ axis. To build the augmented training data, images were rotated with random values from $\{0^\circ, 15^\circ,...,345^\circ\}$ set respect to $y$ axis. For each image, we produced $20$ different images with different random rotations, and each new image with its rotation information was saved in data set. Therefore, the training data contains $500*20=10000$ palmprint images. To have a much generalized data set, other augmentation techniques such shrinkage, adding noise and flipping respect to $y$ axis were applied.

CNN output is  vector $Y$ with length $24$ showing the probabilities of each $24$ classes. The class with highest probability shows the rotation of input palmprint image respect to $y$ axis. For instance, if the probability of class $i, i=0,1,...,23$ is maximum, then the rotation around $z$ axis and respect to $y$ axis is $(i\times15) ^\circ$. To use CNN and find the rotation, for a given input image we smooth the output vector $Y$ by a window with length $3$ and call it $Y_{s}$. Then, maximum value of vector   $Y_s$ is considered, $M_s$ . If  $M_s > 0.35$, then the output of network is reliable. Those images that satisfies this condition, are rotated around $z$ axis by $-(i*15)^\circ$ (correcting rotation, see figure \ref{fig_work_flow}) and are passed to the second round of registration. On the other hand, images with smaller value of  $M_s$ are passed directly to feature extraction step; more likely registration will fail if we pass these images to  ``registration refining". For these palm images we do not apply registration, and treat them differently in local matching step (section \ref{LM}).

\begin{figure}
    \centering
    \includegraphics[width=3.5in]{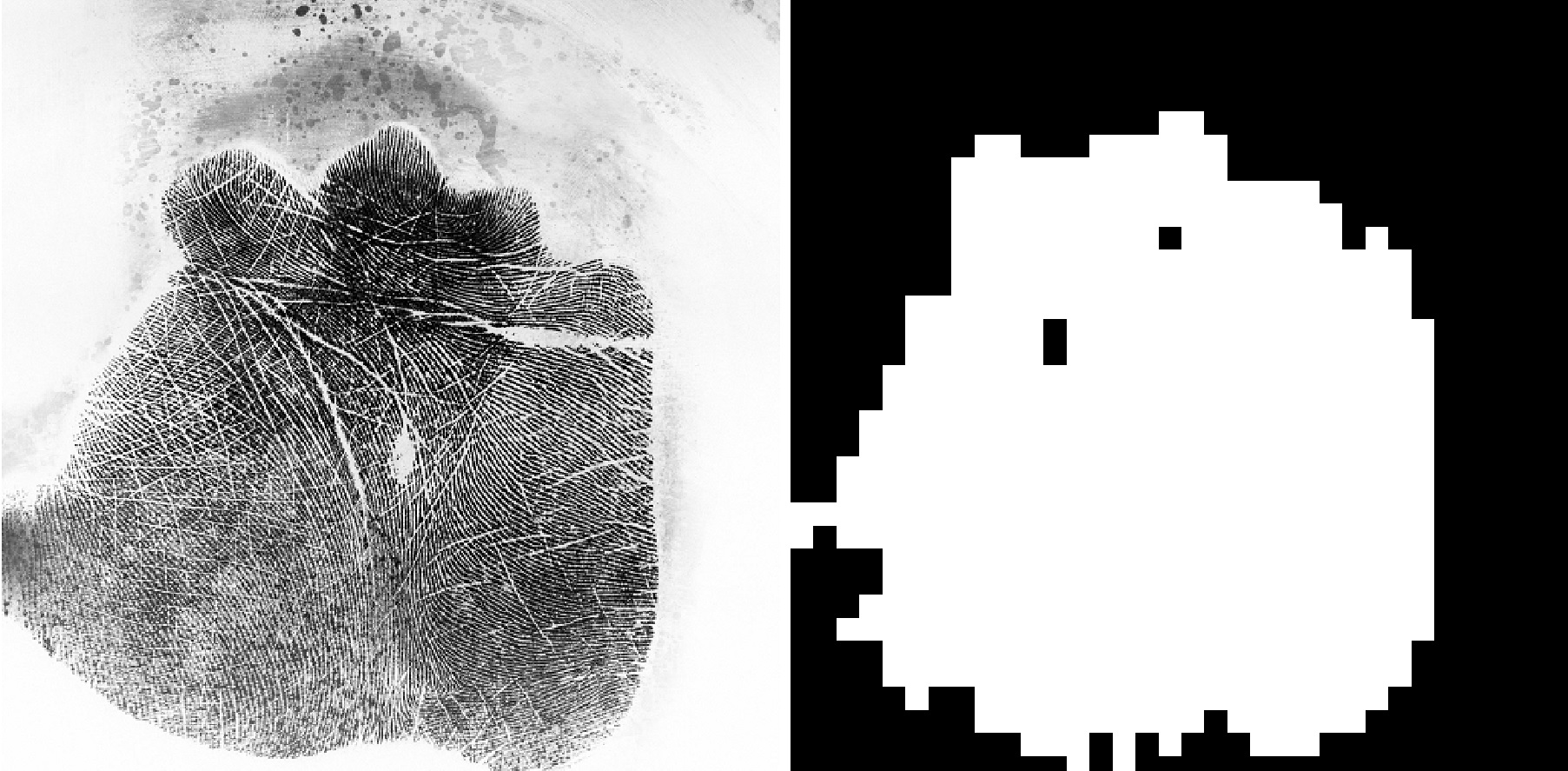}
    \caption{Mask image produced from outputs of layer \#3 of designed CNN}
    \label{fig_mask}
\end{figure}

Another advantage of designed CNN is its ability to segment palm region  from  the background.  We noticed that  in trained model, the outputs of layer \#3 are  near binary image which carry high values for regions with palm, and they near to zero for pixels belonging to the the background. In other word, during the training phase, the coefficient of filters of this layer were tuned in a way that they give the segmented palmprint image. Figure \ref{fig_mask} illustrates the resulted mask image produced by combination of outputs of layer \#3 for a sample palmprint. As it is seen, it can be used to segment the foreground from background. Palmprint segmentation is a pre-step before minutia extraction. Now, we have this image without any extra computation.

\subsection{Registration refining by GHT}\label{GHT_reg}
 In previous section, by assigning each palmprint to one of $24$ classes of angles, we roughly found the rotation difference between palmprint images respect to ground truth. To find exact values of registration (displacement in $x$ and $y$ directions and rotation around $z$ axis), we apply GHT. We know that rotation around $z$ axis can not be larger than ${\frac{360}{24}} = 15^\circ$ (we already corrected the rotation difference, $\theta$ ). This will help us to find the exact value of $\theta$ much faster.
   CNN outputs ($M_s$) are used to judge the level of confidence of registration. If it is larger than a threshold, it means that the estimated $\theta$ is reliable. For  images with small values of $M_s$  we can not trust the estimated $\theta$, and as a result we ignore doing registration on these images (see figure \ref{fig_work_flow}).\par
 
 Similar to  \cite{SoleimaniAhmadi2018}, we used orientation field of each palmprint image to find the registration parameters. The average orientation field of palmprints is used as  the reference image, and other palmprint are aligned with this image. The proposed orientation field estimation algorithm in  \cite{SoleimaniAhmadi2018} is used to compute the orientation field in each $16\times16$ blocks. Finally, the GHT algorithm  \cite{Ballard1981}, is applied to find registration parameters.

All possible pairs of $16\times16$ blocks between the input (unregistered) palmprint orientation and the reference orientation field vote for corresponding rotation and the displacement (in $x$ and $y$ directions). Contrary to \cite{Daietal2012} which uses inverse of circular standard deviation as weights of each block in voting system, we calculate the weight of each block based on its quality \cite{SoleimaniAhmadi2018}.

Palmprint registration,  brings palm images to the same coordinate, which helps to speed up the matching, with a small accuracy reduction. We need a criterion to judge the result of the registration and to check if it can be trusted or not. To measure the reliability of registration results, we define few new parameters:
\begin{itemize}
     \item {$q_1=\frac{\# of \:all \:blocks\: in\: foreground}{\# of \:all \:blocks\: in\: image}$} 
    \item {$q_2=\frac{\# \: blocks\: vote\: for\: best \:candidate\:(\theta, d_x,d_y)}{\# of \:all \:blocks\: in\: foreground}$} 
\end{itemize}

$q_1$ shows the percentage of blocks in an image which belong to palmprint, i.e., foreground.  A larger value of $q_1$ shows the higher quality of a palmprint or it could also indicate larger palmprint image.  Small values of $q_1$ show that the palmprint image is not a full palmprint (latent palmprint) or the palmprint image has low quality. If this value is less than a threshold (in our case $0.3$), we ignore registration for that palmprint. $q_2$ shows the quality of registration, and is the ratio of blocks which vote for final registration parameters to all blocks in foreground. $q_2$ with larger values guarantees the registration accuracy; $q_2 > 0.5$ is acceptable.  These parameters can be used to determine if the input palmprint is for right hand or left hand. 

The reference orientation palmprint which we use for refining the registration , belongs to left hand. GHT is applied to do exact registration for both original and flipped version of the input palmprint (flipped around $y$ axis). Now, two $q_2$s are in hand, resulted from registration of reference image with orientation of input palmprint and its flipped version. If $q_2$ of original palmprint is larger than $q_2$ of flipped image, then input palmprint is left hand and vice versa. Note that, if the larger $q_2$ is less than  $0.5$, then the registration   has most likely failed. Therefore, our algorithm can automatically recognize left palmprint from right palmprint which will increase the palmprint matching speed and accuracy. 
\begin{figure}
    \centering
    \includegraphics[width=3.5in]{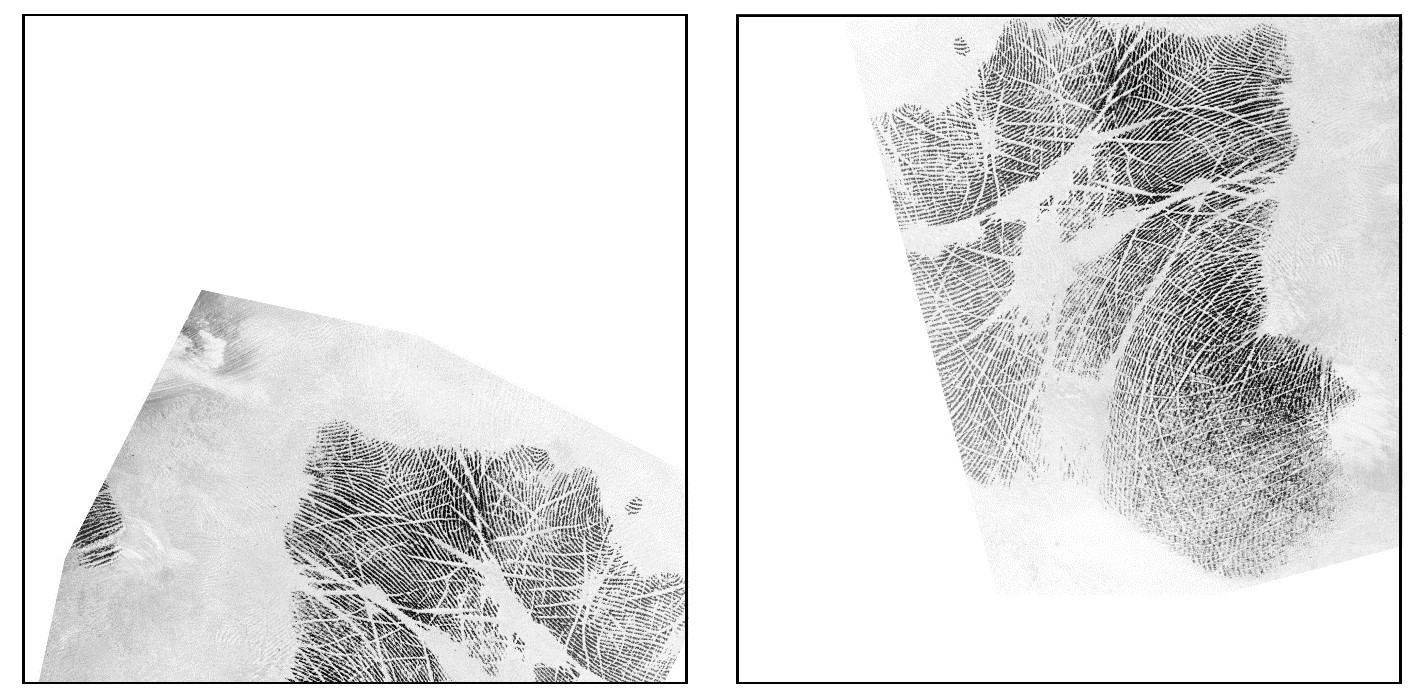}
    \caption{Results of final registration for both original(left) and flipped image(right). For both images $q_1=0.36$, for original image $q_2 = 0.3$ and for flipped version $q_2=0.89$. }
    \label{fig_left_right}
\end{figure}

 \begin{figure*}[bt]
    \centering
    \includegraphics[width=7in]{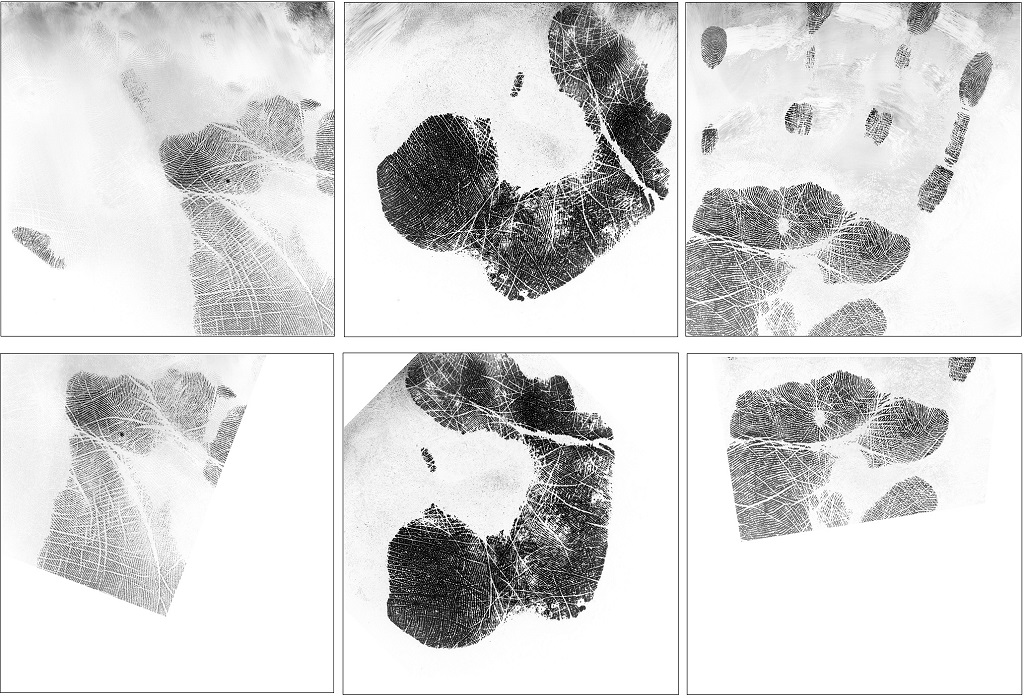}
    \caption{ Sample palmprint registered by proposed method. First row shows the original images and the second raw indicates the registered images }
    \label{fig_sample_register}
\end{figure*}
 
 \begin{figure*}[ht]
    \centering
      \begin{subfigure}[b]{0.3\textwidth}
        \includegraphics[width=\textwidth]{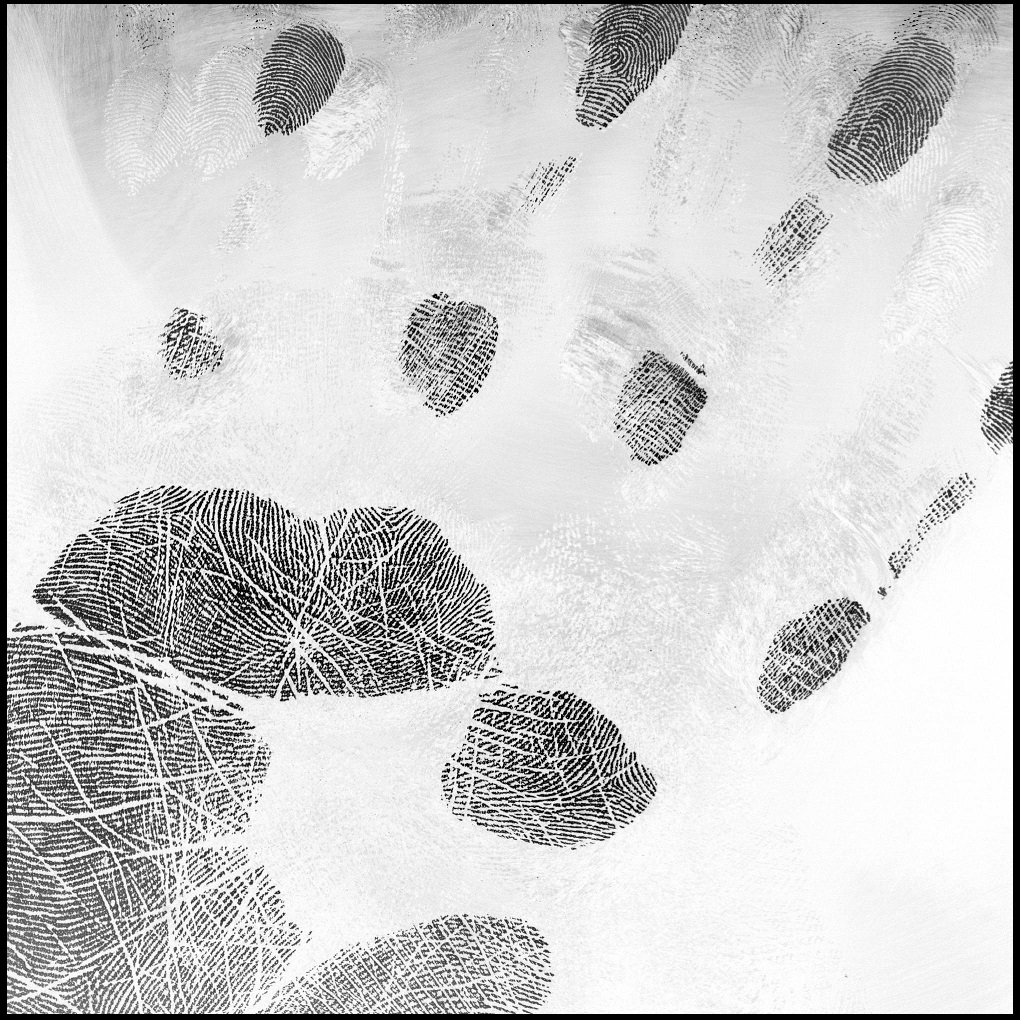}
        \caption{Original image}
    \end{subfigure}
      \begin{subfigure}[b]{0.3\textwidth}
        \includegraphics[width=\textwidth]{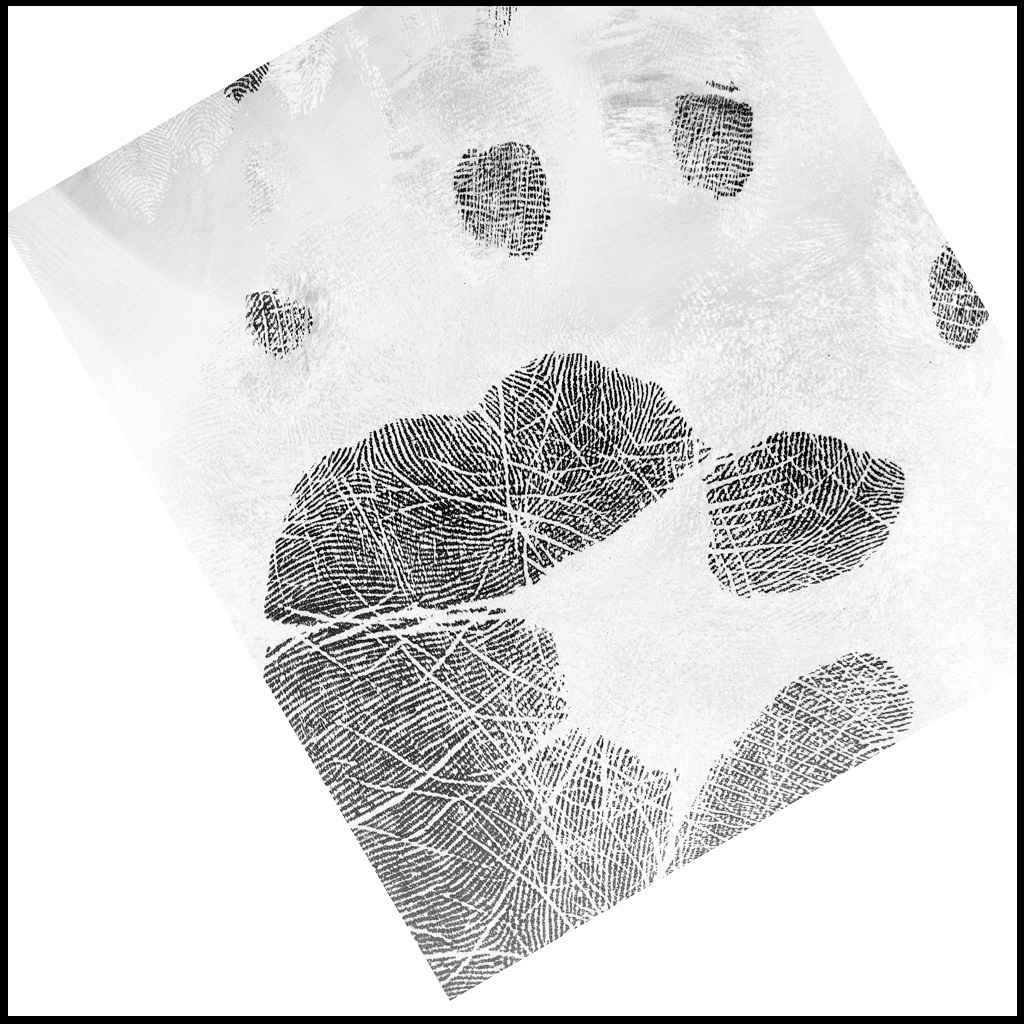}
        \caption{Registered by CNN}
    \end{subfigure}
      \begin{subfigure}[b]{0.3\textwidth}
        \includegraphics[width=\textwidth]{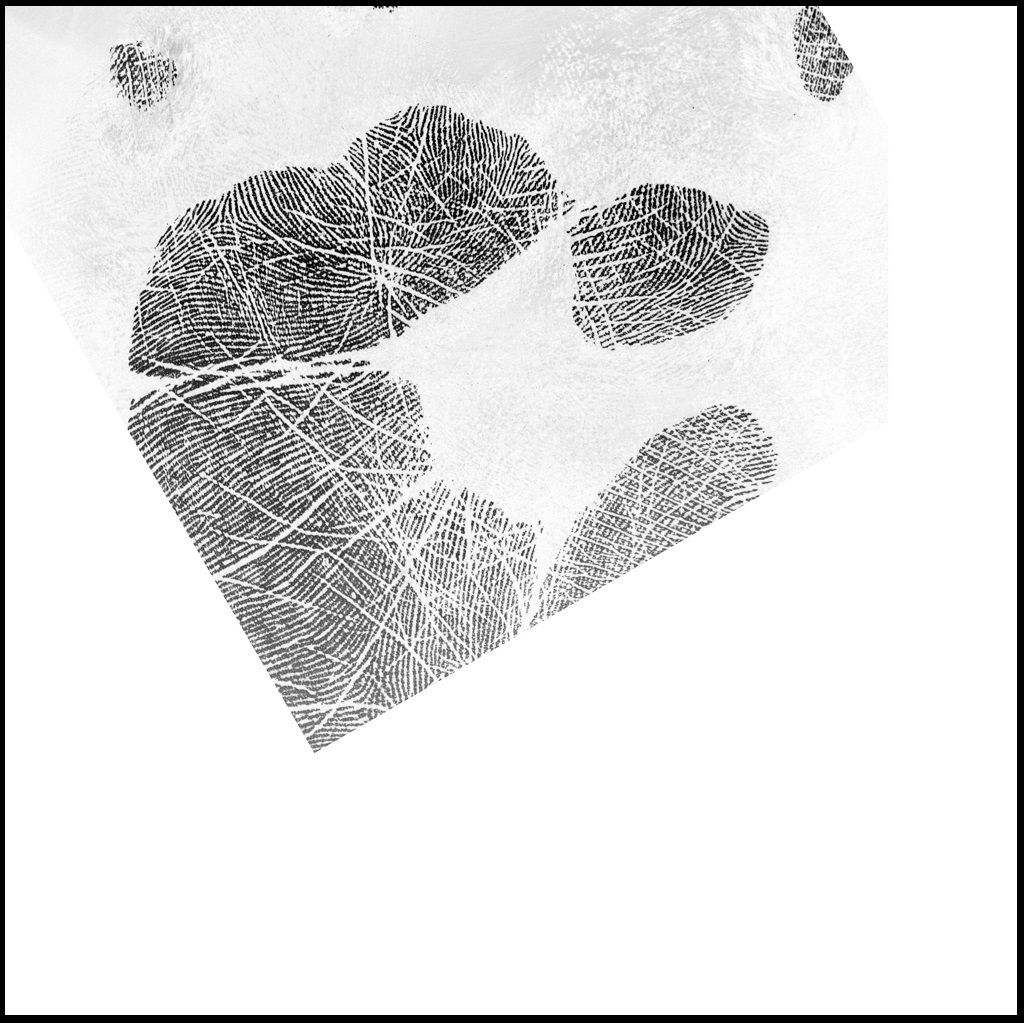}
        \caption{Registered by GHT}
    \end{subfigure}
      \begin{subfigure}[b]{0.3\textwidth}
        \includegraphics[width=\textwidth]{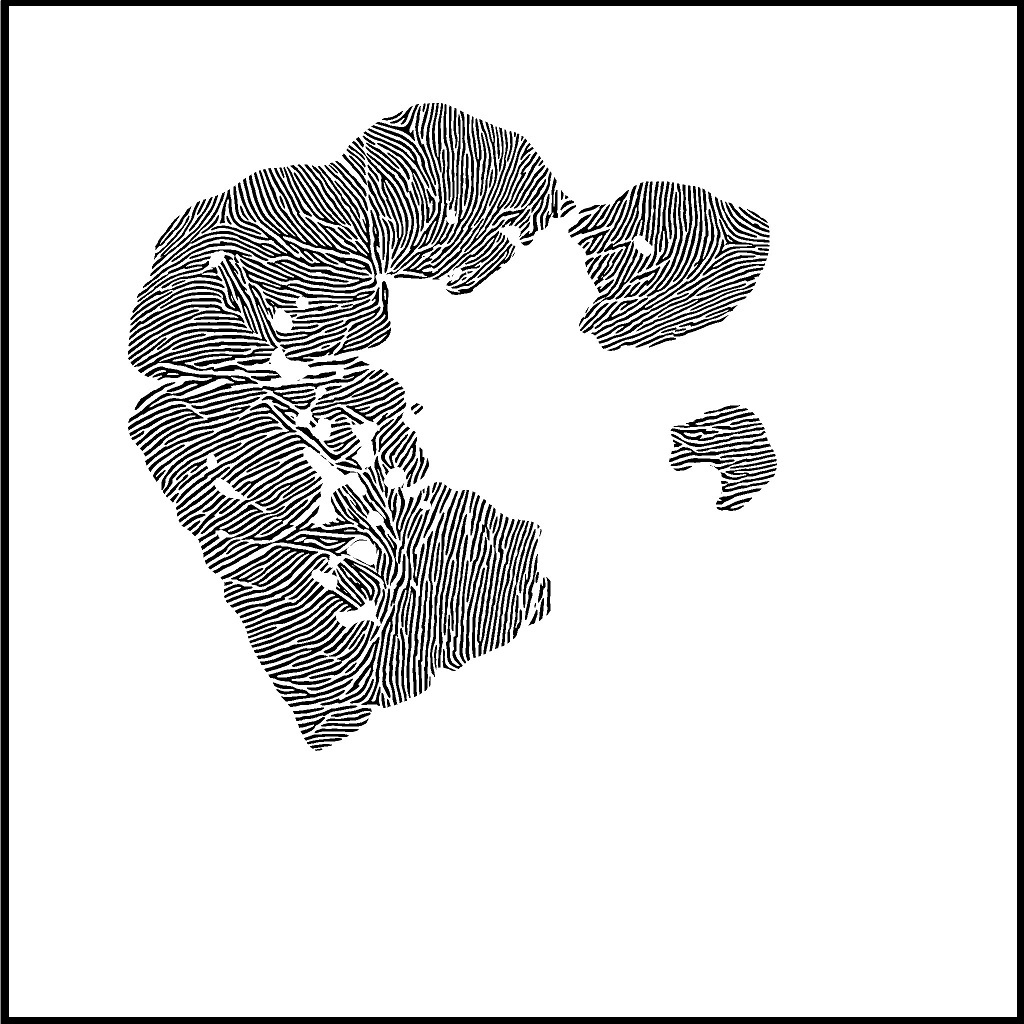}
        \caption{Enhanced image}
    \end{subfigure}
      \begin{subfigure}[b]{0.3\textwidth}
        \includegraphics[width=\textwidth]{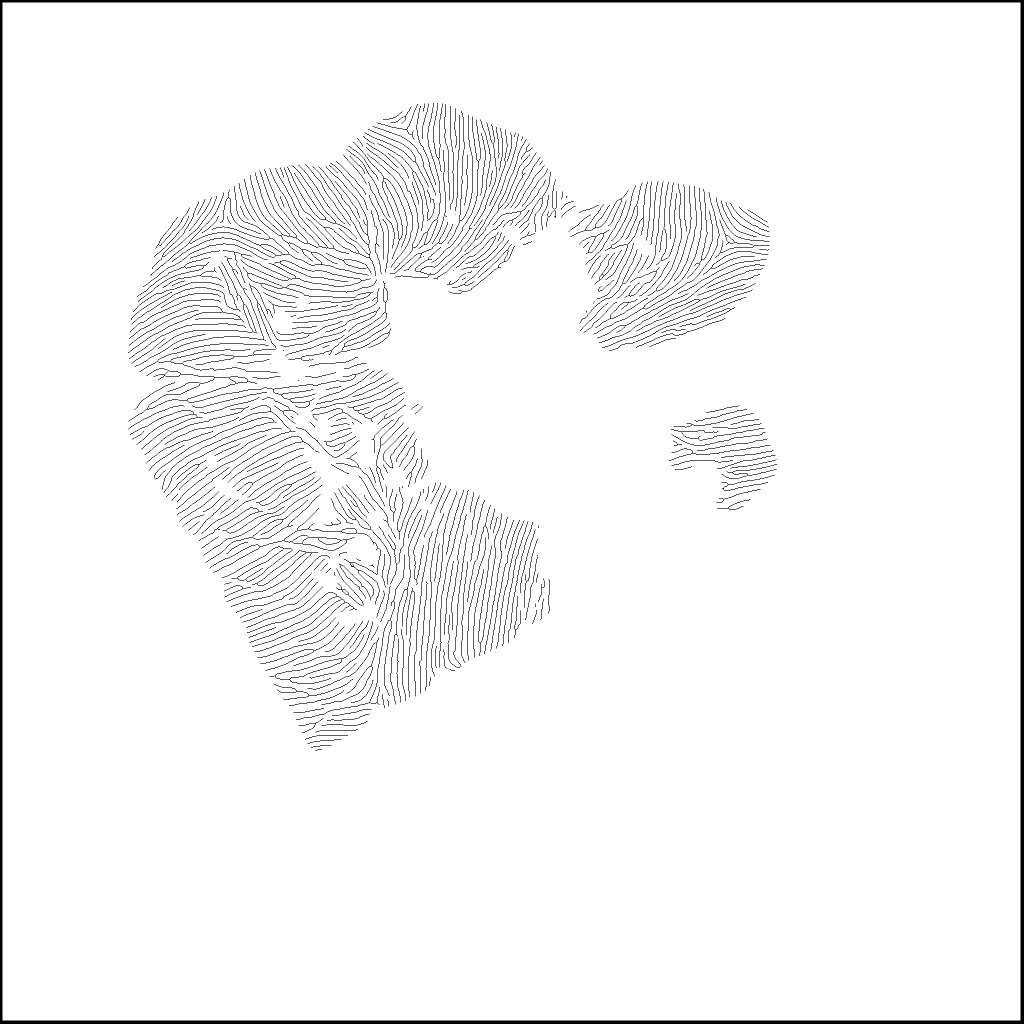}
        \caption{Skeleton image}
    \end{subfigure}
      \begin{subfigure}[b]{0.3\textwidth}
        \includegraphics[width=\textwidth]{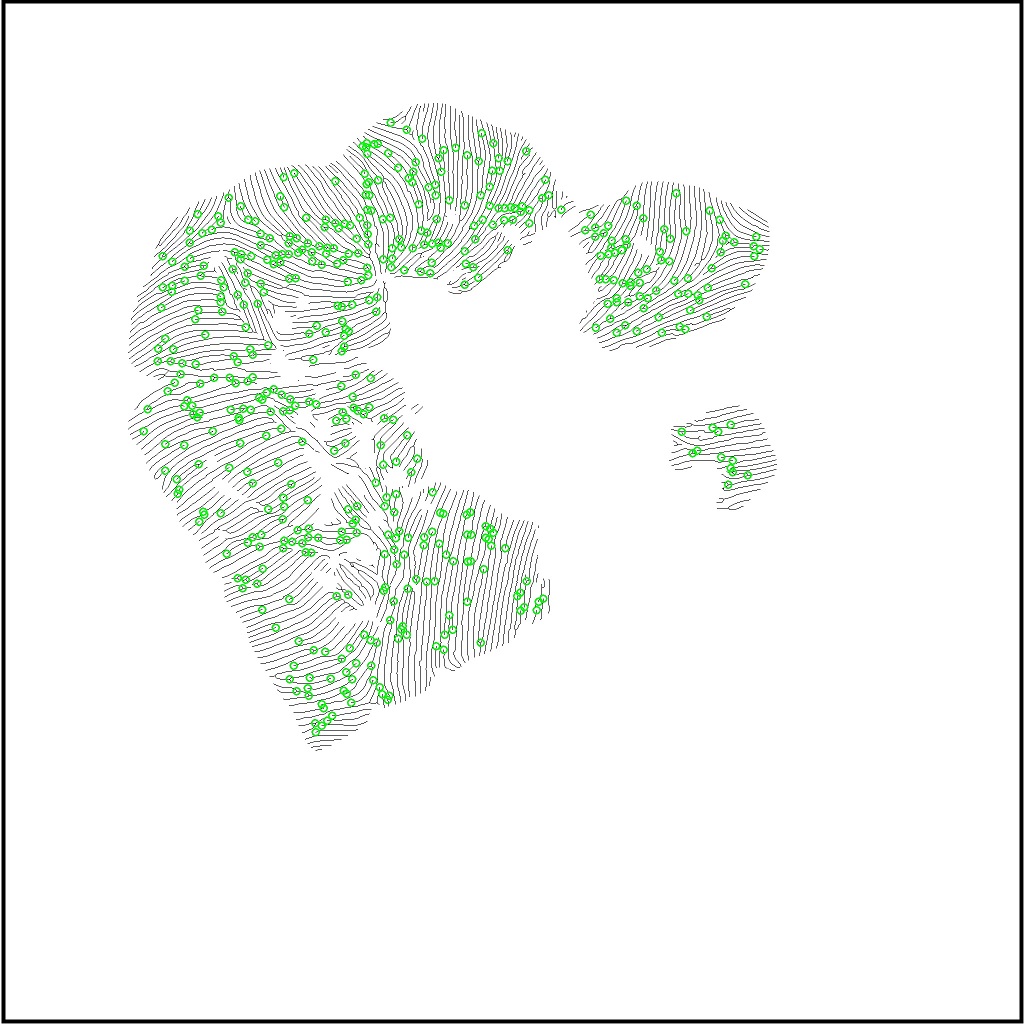}
        \caption{Minutiae}
    \end{subfigure}

    \caption{Sample output of each step of whole procedure }
    \label{steps_of_whole_process}
\end{figure*}

Figure \ref{fig_left_right} shows results of registration for a sample right hand palmprint. $q_1=0.36$ satisfies the threshold while $q_2$ for left hand palmprint is less than $0.5$ and it is much smaller than $q_2=0.89$ of right hand palmprint. Therefore, features (Minutia) are extracted from right image, and palmprint class is known, i.e., the palmprint belongs to right hand.

We applied the proposed registration strategy to THUPALMLAB database(train set) and only a few of the images did not satisfy  $q_2 > 0.5$.  Maximum allowed rotation was limited to $30 ^ \circ$  (with steps=$3 ^ \circ$) and maximum displacement was restricted to $500$ pixels, which is  $\frac{1}{4}$ of the image width and height. For palmprints that $q_1$ or $q_2$ do not satisfy thresholds, palmprint class is ``unknown".

 Figure \ref{fig_sample_register} shows some raw palmprint samples, from the same hand, and their registered version. As it is seen in the second row, all three images are almost in the same coordinate.

\section{Palmprint Matching}\label{match}
 After registration, the segmented palmprint is enhanced by applying Gabor filters \cite{Hongetal1998}. The enhancement, which is guided by the estimated local orientations  and frequencies, produces a near binary image which is then simply binarized. The thinning algorithm in \cite{ZhangSuen1984} is applied to the enhanced binary image and the palmprint skeleton is obtained. Using the algorithm in \cite{Maltonietal2009} minutiae are extracted and the results are refined by approach in \cite{ZhaoTang2007}.  Once the minutiae have been extracted, the matching process can be done. Figure \ref{steps_of_whole_process} shows the  whole procedure for sample palmprint image.\\

The matching algorithm is similar to what we have done in \cite{SoleimaniAhmadi2018}. It consists of two stages: {\bf a}) local minutia matching and {\bf b}) global score calculation. In the local minutia matching stage, the similarity of a minutia pairs is computed with the MCC descriptor to make the matching process more efficient and fast. In the second stage, overall similarity of two palmprints is calculated.\\

\subsection{Local matching}\label{LM}

The MCC descriptor is a local data structure which is invariant
for translation and rotation. This structure encodes spatial and directional relationships among the minutia and its (fixed-radius)
neighborhood. It is represented by a bit
vector of fixed length. 
To check the similarity of two local minutiae from two different images, we  compare their MCC descriptors. For $N_S =16$ (number of cells along the cylinder diameter) and $N_D=5$ (number of cylinder section), each MCC descriptor contains a vector with length  equal to $16\times16\times5=1280$ bits. To check the similarity of two minutiae, we need to do a $XOR$ operation between two vectors, with length $1280$, which is a time consuming process and may not be suitable for palmprint images. To decrease the number of  similarity computation, we use registered images. In other word, we only check the MCC similarity between two minutia which roughly belong to the same part of the palm. By decreasing the number of MCC similarity computations, the matching process will speed up. For a given minutia $m_A(x_{mA},y_{mA},\theta_{mA})$ from palmprint A and minutia $m_B(x_{mB},y_{mB},\theta_{mB})$ from palmprint B; If $m_A$ and $m_B$ do not belong to the same region, their local similarity is zero. This means, if $|x_{mA}-x_{mB}| > T_x$, $|y_{mA}-y_{mB}| > T_y $ or $\theta_{mA}-\theta_{mB} > T_{rot}$ their MCC similarity is set to zero. For images which fail to be registered by CNN, $T_{rot}=180 ^ \circ, T_x=T_y=2048$. Images with small values of $q_1$ or $q_2$ in second phase of registration(refining by GHT), $T_{rot}$,$T_x$ and $T_y$ are set to $30^\circ$, $1600$ and $1600$ respectively. Finally, for images which are successfully registered, $T_{rot}=20^\circ, T_x=T_y=500$ (see figure \ref{fig_work_flow}).

Without registration , the original MCC algorithm needs to compare $N_A$ minutiae from image $A$ with $N_B$ minutiae from image $B$, where $N_A$ and $N_B$ are number of minutia in image $A$ and $B$, respectively. This amount of comparison is time consuming and use of  registration  decrease this complexity, dramatically.

\subsection{Global matching}\label{GM}
In order to compare two palmprints, a global score (denoting their overall similarity) needs to be derived from the local similarities. The  approach proposed in \cite{SoleimaniAhmadi2018} is used to calculate the global
score.
For two given  MCC minutia descriptor sets ${\bf A}=\lbrace {MCC_{a_1}, MCC_{a_2},...,MCC_{a_{N_A}}} \rbrace$ and ${\bf B}=\lbrace {MCC_{b_1}, MCC_{b_2},...,MCC_{b_{N_B}}} \rbrace$, there is a local similarity score matrix,   $S_{N_A\times N_B}$, showing the MCC similarity between minutiae of image $A$ with those of image $B$. $S(i,j)$ is the MCC similarity of minutia $i$ from image $A$ with minutia $j$ from image $B$. Based on the section \ref{LM}, $S$ is a low density matrix; most of its elements is zero. \par
To compute the global score, $n_R$ minutia pairs with maximum similarity values are selected. Suppose $P = \lbrace(r_t, c_t), 0 < r_t < N_A\:\: and \:\:0 < c_t < N_B \rbrace$ is the selected $n_R$ minutiae-index pairs. First $\hat{S}$, normalized version of $S$, is computed:

\begin{equation}
\hat{S}(i,j)=\langle  1- {{\sum^{N_A}_{\begin{array}{c}
k=0 \\
k\neq i 
\end{array}
}
S(k,j)  +
\sum^{N_B}_{\begin{array}{c}
k=0 \\
k\neq j 
\end{array}}
S(i,k) }\over{N_A+N_B-2}}\rangle S(i,j).
\end{equation}
 
 The normalization  modifies each value according to
the average of the values in the same row and in the same column.  
After  normalization, $n_R$ minutiae pairs, corresponding to the top
$n_R$  values in $\hat{S}$ matrix, are selected and  a relaxation procedure  is
applied to the $n_R$ similarity scores guided by their corresponding minutia pairs.\par
 Let $\hat{s}^0_t=\hat{S}(r_t,c_t)$ is initial normalized similarity of pair $t$; the similarity of pair $t$ at iteration $i$ of the relaxation procedure is:

\begin{equation}
\label{relaxation_eq}
\hat{s}^{i}_t=w_R. \hat{s}^{i-1}_t +(1-w_R).{{\sum^{n_R}_{k=1, k\neq t}\rho(t,k).\hat{s}^{i-1}_k}\over{n_R-1}},
\end{equation}
where $w_R \in[0, 1]$ is a weighting factor and:\\

$\qquad \rho(t,k)=\prod_{i=1}^{3} Z(d_i,\mu^{\rho}_{i}, \tau^{\rho}_{i}  ),$
\begin{equation}
\begin{array}{c}
\! d_1={{|d_s(a_{r_t},a_{r_k})- d_s(b_{c_t},b_{c_k})|}\over{max(d_s(a_{r_t},a_{r_k}), d_s(b_{c_t},b_{c_k}))}},\\
\quad d_2=|d_\varphi(d_\theta(a_{r_t},a_{r_k}),d_\theta(b_{c_t},b_{c_k}))|,\\
\quad d_3=|d_\varphi(d_R(a_{r_t},a_{r_k}),d_R(b_{c_t},b_{c_k}))|,
\end{array}
\end{equation}
where $d_s(m_1,m_2)$ shows the  Euclidean distance between minutia $m_1$, $m_2$ and $d_\varphi(\theta_1,\theta_2)$ is the difference (modulo $2\pi$) between two angles $\theta_1$, $\theta2$. $d_\theta(m_1,m_2)$ indicates directional difference between two minutiae ($d_\theta(m_1,m_2)=d_\varphi(\theta_1,\theta_2)$). $\rho(t,k)$ measures compatibility between two minutiae pairs $(a_{r_t},a_{r_k})$ from image A and $(b_{c_t},b_{c_k})$ from image B. It is computed as the product of three feature, $d_1, d_2$ and $d_3$ and normalized by a sigmoid functions with its specific $\mu$ and $\tau$:
\begin{equation}
Z(\nu,\mu,\tau)={{1}\over{1+e^{-\tau(\nu-\mu)}}},
\end{equation}
$d_1$ computes distance similarity of two minutia pairs. 
$d_2$ is the similarity of directional differences and $d_3$ denotes the radial angle  similarity. Radial angle is the angle between line connecting two minutiae and the direction of the first minutia. The smaller $d_1$, $d_2$ and $d_3$ results in a smaller similarity value between two minutia pairs $\lbrace (a_{r_t},a_{r_k}), (b_{c_t},b_{c_k})\rbrace$.

Relaxation procedure (\ref{relaxation_eq}) is repeated for $n_{rel}$ times and an efficiency of each pair $t$ is computed as follow:
\begin{equation}
\varepsilon_t={{\hat{s}^{n_{rel}}_t}\over{\hat{s}^0_t}}.
\end{equation}

At the end, global score of two palmprint images A and B is calculated as the average of the $n_P$ similarity values, $\hat{s}^{n_{rel}}_t$, corresponding to top $n_P$ pairs with highest efficiency values.


\section{Experimental results}\label{exres}
In this section, experiments were carried out to evaluate the proposed method. The proposed method is  compared  with other palmprint matching algorithms, and results and parameters are reported.\par

\subsection{Database and Parameter Tuning}\label{parametr_tuning}
The only publicly available high-resolution palmprint database, Tsinghua Palmprint Database \cite{Dai2012}, is used as the main database. It consists of $1280$ palmprint images from
$80$ different people (left and right palms and eight impressions per palm). All   images are $2040 \times 2040$ pixels with  $500$ dpi resolution and  256 gray levels. By zero padding, the image size is changed to $2048\times2048$ ($2048$ is divisible by $16$ which is block-size in orientation field estimation algorithm). The THUPALMLAB database consists of two separate sets: 1) the training set
and 2) the test set. The training set (the first $20$ subjects) includes $320$ palmprints (from $40$ different palms), while the test set contains the remaining $960$ palmprints (from $120$ different palms).The training set is used to tune the parameters, and it is done by minimizing  Equal Error Rate (EER). \par

Thanks to the registration, the MCC similarity of  small percent of all possible minutia pairs  is non-zero, i.e, matrix $S$ is sparse.  In the global score computation phase, $n_R$ minutia pairs with maximum local similarity score are selected from $\hat S$ Matrix. Therefore, the remaining  non-zero elements of the similarity matrix $S$ are more reliable, and the relaxation phase can be done even  with small values of $n_R$.  Reducing $n_R^{max}$ from $300$(in original MCC) to $70$ decreases the relaxation phase computations dramatically. With $n_R^{max}=300$, $\rho(t,k)$ needs to be calculated almost $300\times300=90000$ times in each iteration and this value is much bigger than $70\times70=4900$. In contrast to \cite{Cappellietal2012} where $N_S = 16$,  in our proposed system, $N_S$ is $8$ and the resulted MCC descriptor length is $8\times8\times5=320$ bits, and it takes much less time to compute the MCC similarity of two minutiae. Also, since palmprint images are registered, the maximum global rotation between two images cannot be a large value.\par

\subsection{Time Complexity and Computational Requirements}

A typical palmprint  contains $1000$ minutiae on average; then the original MCC algorithm needs to do $1000\times1000=10^6$ MCC local similarity computations.
Our statistical observation showed that by applying proposed registration algorithm, we only need to do $12$ percent of all computations, i.e. $1.2\times10^5$ local MCC similarity computation. Results in table \ref{Time_table} confirm this, and show that the contribution of the registration  toward the matching time reduction is significant.  By the proposed system, ${{1000}\over{9}}=111$ palmprint matches per second can be done whereas the feasible number of matches per second for algorithm in \cite{Cappellietal2012} (the  fastest algorithm to the best of our knowledge) is  ${{1000}\over{36}}=26$. Note that, we can double the matching speed by applying proposed left-right palm detector, i.e. average  matching time is $4.5$ milliseconds(222 matches per seconds). \par

\begin{table}
\centering
\caption{Average matching time (second)}

 \begin{tabular}{|p{6cm}|c|}
  \hline
  
  \textbf{Method} & \textbf{Matching time}\\
  \hline
Jian and Feng \cite{JainFeng2008}& $1.097 $ \\ \hline
Dai and Zhou \cite{DaiZhou2011}& $5.160$ \\ \hline
Chen and Guo \cite{ChenGuo2016}& $0.089 $ \\ \hline
Tariq et.al(CPU) \cite{tariq2017massively}& $0.031 $ \\ \hline
Proposed method & {\bf 0.021} \\ \hline

Capelli and Ferrara \cite{Cappellietal2012} (Multithreaded)& {$0.038$} \\ \hline
Proposed method (Multithreaded) & {\bf 0.006} \\ \hline
   
 \end{tabular}
\label{Time_table}
\end{table}

\begin{table}
\centering
\caption{Required memory for a typical palmprint image}

 \begin{tabular}{|c|c|}
  \hline
  
  \textbf{Method} & \textbf{Required memory}\\
  \hline

Jian and Feng \cite{JainFeng2008}& $240 KB$ \\ \hline
Dai and Zhou \cite{DaiZhou2011}& $4084 KB $ \\ \hline
 Chen and Guo \cite{ChenGuo2016}& {\bf  less than 8 KB} \\ \hline
 Tariq et.al \cite{tariq2017massively}& {  $18 KB$} \\ \hline
Capelli and Ferrara \cite{Cappellietal2012}& {$161 KB$} \\ \hline

Proposed algorithm& { $45 KB$} \\ \hline
 
 \end{tabular}
\label{required_memoty}
\end{table}

\begin{table*}[!ht]
\caption{EER and FMMR at give FMRs} 
\label{Result_table}
{\begin{tabular*}{\textwidth}{@{\extracolsep{\fill}}c|cccc}
  
  \textbf{Method} & \textbf{EER} & \textbf{FNMR at FMR$\leq.01\%$}
  &  \textbf{FNMR at FMR$\leq.001\%$} & \textbf{FNMR at FMR $ =0$} \\
\hline
Jian and Feng \cite{JainFeng2008}& $5.04 \%$ & $17.32\%$ & $19.43 \%$ & $22.5\%$\\
\hline

Dai and Zhou \cite{DaiZhou2011}& $2.99 \%$ & $8.78\%$ & $10.45 \%$ & $11.58\%$\\
\hline

Wang and Ram \cite{Wangetal2014}& $1.77 \%$ & $> 10\%$ & $ > 10 \%$ & $>10\%$\\
\hline

Chen and Guo \cite{ChenGuo2016}& $0.29 \%$ & $2 \%$ & $5 \%$ & $ > 10\%$\\
\hline

Tariq et.al \cite{tariq2017massively}& $0.38 \%$ & $ > 0.38\%$ & $ > 0.38\%$ & $ > 0.38\%$\\
\hline

Capeli and Ferara \cite{Cappellietal2012}& {\bf0.01\%} & {\bf 0.1\%} & $0.18 \%$ & $0.48\%$\\
\hline  
Proposed method& {$0.04\%$} &  $0.12\%$ & {\bf0.14\%} &{\bf0.24\%}\\

\end{tabular*}}{}
\end{table*}

The proposed algorithm was tested on an Intel(R) Core(TM) 2 Quad Q9550
CPU at 2.83 GHz, using  single and multi-threaded  C++ implementation. Table \ref{Time_table} summarizes average  matching times of the proposed system with those of four other approaches in \cite{JainFeng2008}, \cite{DaiZhou2011}, \cite{ChenGuo2016},\cite{tariq2017massively} and \cite{Cappellietal2012}.
 The times of the methods in \cite{JainFeng2008}, \cite{DaiZhou2011} have been
measured on an Intel Xeon E5620 at 2.4 GHz, with a single-thread in
C++, and the time of \cite{ChenGuo2016} has been measured by Intel(R) core (TM) i7 CPU
2.8 GHz in C++. Time of the algorithm in \cite{Cappellietal2012} has been measured by Intel Core 2 Quad Q9400 CPU at 2.66 GHz, using a multi-threaded C\# implementation. Due
to  different hardware and the way that software is implemented, it is hard to accurately  compare the matching times of the algorithms. For example, we implemented the method in \cite{Cappellietal2012} in our system using a multi-threaded C++ implementation; and the matching process took about $200$ milliseconds, while the authors of this algorithm report a matching time of $38$ milliseconds with a weaker computer and C\# implementation. 
 
 The computational cost of the registration procedure
 is composed of applying CNN,voting and  searching for the best transformation parameters. The execution time of the registration 
is about four seconds in our experiment. Since it is performed
in the enrollment stage only once, it has  no computation complexity during  identification. \par

 The amount of memory which is required by the proposed algorithm to store
the palmprint templates is relatively small in comparison to other algorithms.
 For a single minutia, the proposed system stores  $8\times8\times5$ bits for the MCC descriptor, and about five bytes for ($x_m, y_m, \theta_m$). In total it needs to store $40+5 = 45$ bytes per minutia which is a small value in comparison to the $161$ bytes (per minutia) required by the algorithm in \cite{Cappellietal2012}. The method in \cite{LiShi2008} requires $230$ bytes per minutia and one byte per orientation element; the algorithm in \cite{Mangold2016} needs five bytes per minutia, two bytes per orientation.  For a typical palmprint ($2040 \times 2040$) with an average $1000$  minutiae and  $127\times 127$ orientation elements, the amount of memory required by these algorithms is summarized in table \ref{required_memoty}. Among all these algorithms, the proposed system requires relatively less amount of memory to store minutia descriptors and templates. The method in \cite{ChenGuo2016} needs less memory in comparison to ours, but the matching time and the accuracy of this algorithm are much worse.

\subsection{ Accuracy }

 To evaluate the matching algorithms, impostor and genuine tests need to be done. For this database,
the total number of genuine matches is $({{8\times7}\over{2}}\times120=3,360)$. Each impression is compared with the remaining impressions of the same palmprint. The number of impostor matches is much larger than that of genuine matches. All $8$ impressions from $120$ subjects are compared with the remaining impressions; therefore, the number of impostor matches is $({{952\times960}\over{2}}=456,906)$.\par

The accuracy of the proposed system is compared to the some state-of-the art(
\cite{JainFeng2008}, \cite{DaiZhou2011}, \cite{Cappellietal2012},\cite{Wangetal2014} and \cite{ChenGuo2016}). As  seen in Table \ref{Result_table}, our algorithm  shows a  higher accuracy at any FMR. At zero FMR, the proposed algorithm makes $8$ false non-match errors. Most of these errors occur because of the images with low quality, regions with small area and less pattern (similar to latent palmprints), and less overlap area with other palmprints. When two palmprint images have no minutia in common, i.e. no overlap, non of the existing algorithms can find a high matching score between them. 
It is noteworthy that without registration , the accuracy of the proposed system reduces dramatically. This is because of the small values of $n_R$, $N_S$  and some other parameters that we have changed in MCC algorithm.\par
 
As for the accuracy of the registration phase, $5$ palmprints were not successfully registered, which is less than one percent of all the palmprints within the test database. All of the failure cases are due to  bad image quality or  pattern-less and small areas. In all of these $5$ images, one of $M_s, q_1$ or $q_2$ was less than the corresponding threshold. 


\section{Conclusion}\label{conc}
Because of the existence of creases, a  large number of minutia and skin distortion, minutia-based palmprint matching is a challenging task. In this paper, we  designed new strategy to speed up the palmprint matching speed. 
Combining CNN and GHT, an accurate palmprint registration was designed which can  bring all palmprints to a same coordinate system. In the first round of registration, CNN finds the rotation $\theta$ of a palmprint image respect to $y$ axis. Image is rotated by $(-\theta)$, and exact registration is done by GHT. Registration helps to compare only minutiae belonging to the same part of the palmprint, ignoring comparison of minutiae which are far apart. This reduces computations in local match phase, and makes the matching algorithm faster.
 
Our method  is six times faster than the state-of-the-art while the matching accuracy is still very high.The proposed algorithm was applied on THUPALMLAB database and the results revealed that  our algorithm has the capability of doing  $166$ palmprint matches per second with $EER = 0.04 \%$.


\bibliographystyle{IEEEtran}
\bibliography{paper}

\end{document}